\documentclass[runningheads]{llncs}

\usepackage{caption} 

\usepackage[hyphens]{url} 
\usepackage{graphicx} 
\usepackage{hyperref}
\usepackage{url}            
\usepackage{amsfonts}       
\usepackage{nicefrac}       
\usepackage{microtype}      
\usepackage{graphicx}
\usepackage{amsmath,amssymb}
\usepackage{bm}
\usepackage{array,booktabs,ragged2e}
\usepackage{cleveref}
\usepackage{subfig}

\usepackage{multicol}
\usepackage[ruled,vlined]{algorithm2e}

\begin{document}

\title{Neural Clinical Event Sequence Prediction through Personalized Online Adaptive Learning}

\titlerunning{Neural Clinical Event Prediction through Personalized Adaptive Learning}

\vspace{-9mm}

\author{Jeong Min Lee\orcidID{0000-0001-8630-0546} \and Milos Hauskrecht\orcidID{0000-0002-7818-0633}}
\institute{Department of Computer Science, University of Pittsburgh, Pittsburgh, PA, USA
\email{\{jlee, milos\}@cs.pitt.edu}}


\maketitle

\vspace{-8mm}

\begin{abstract} 
Clinical event sequences consist of thousands of clinical events that represent records of patient care in time. Developing accurate prediction models for such sequences is of a great importance for defining representations of a patient state and for improving patient care. One important challenge of learning a good predictive model of clinical sequences is patient-specific variability. Based on underlying clinical complications, each patient’s sequence may consist of different sets of clinical events. However, population-based models learned from such sequences may not accurately predict patient-specific dynamics of event sequences. To address the problem, we develop a new adaptive event sequence prediction framework that learns to adjust its prediction for individual patients through an online model update. 
\end{abstract}

\vspace{-10mm}

\section{Introduction}
\vspace{-2mm}
Clinical event sequence data based on Electronic Health Records (EHRs) consist of thousands of clinical events representing records of patient condition and its management, such as administration of medications, records of lab tests and their results, and various physiological signals. Developing accurate temporal prediction models for such sequences is extremely important for understanding the dynamics of the disease and patient condition under different interventions and detection of unusual patient-management actions, and it may ultimately lead to improved patient care \cite{hauskrecht2016outlier}.
One important challenge of learning good predictive models for clinical sequences is patient-specific variability. Depending on the underlying clinical condition specific to a patient combined with multiple different management options one can choose and apply in patient care, the event patterns may vary from patient to patient. Unfortunately, many modern event prediction models and assumptions incorporated into training of such models may prevent one from accurately representing such a variability. The main challenge, which is also the main topic of this paper, is how to recover at least some of the patient-specific behavior of such models. 

We study this critical challenge in context of neural autoregressive models. Briefly, neural temporal models based on RNN, LSTM, and attention mechanism have been widely used to build models for predicting clinical event time-series \cite{choi2016retain,lee2019context,lee_clinical_2020,lee2020multi,lee2021modeling,liu2019nonparametric}. 
However, when built from complex multivariate clinical event sequences, aforementioned neural models may fail to accurately model patient-specific variability due to their limited ability to represent distributions of dynamic event trajectories. Briefly, the parameters of neural temporal models are learned from many patients data through Stochastic Gradient Descent (SGD) and are shared across all types of patient sequences. 
Hence, the population-based models tend to average out patient-specific patterns and trajectories in the training sequences. Consequently, they are unable to predict all aspects of patient-specific dynamics of event sequences and their patterns accurately.  

To address the above problem, we propose, develop, and study two novel event time-series prediction solutions that attempt to adjust the predictions for individual patients through an online model update. First, starting from the population model trained on a broad population of patients, we adapt (personalize) the model to individual patients to better fit patient-specific relations and predictions based on the current history of observations made for that patient. We refer to this model as the patient-specific model. However, one concern with the patient-specific model and adaptation is that it may lose some flexibility by being fit too tightly to the specific patient and its recent condition. To address this, we also investigate a model switching approach that learns how to adaptively switch among multiple prediction models that may consist of both population and patient-specific models. These solutions extend RNN based multivariate sequence prediction to support personalized clinical event sequence prediction. We demonstrate the effectiveness of both solutions on clinical event sequences derived from real-world EHRs data from MIMIC-3 Database \cite{johnson2016mimic}. 

\vspace{-4mm}

\section{Related Work} 
\vspace{-3mm}
\textbf{Patient-specific Models.} 
The problem of fitting patient-related outcomes and decisions as close as possible to the target individual has been an important topic of biomedical research and personalized medicine.
One classic approach identifies a small set of traits or features that help to define a subpopulation the patient belongs to and applies a model built specifically for that subpopulation \cite{huang2015medical,huang2013similarity}. More flexible patient-specific models \cite{fojo2017precision,rizopoulos2011dynamic,visweswaran2005instance}
identify the subpopulation of patients relevant to the target patient by using a patient similarity measure, and then build and apply the model online when the prediction is needed. 

\noindent \textbf{Online Adaptation Methods.} However, in many sequential prediction scenarios, the models that are applied to the same patient more than once create an opportunity to adapt and improve the prediction from its past experiences and predictions. This online adaptation lets one to improve the patient-specific models and their prediction in time gradually. The standard statistical approach can implement the adaptation process using the Bayesian framework where population-based parameter priors combined with the history of observations and outcomes for the target patient are used to define parameter posteriors \cite{berzuini1992bayesian}. Alternative approaches for online adaptation developed in literature 
use simpler residual models \cite{liu2016learning_a} that learn the difference (residuals) between the past predictions made by population models and observed outcomes on the current patient. Liu and Hauskrecht \cite{liu2016learning_a} learn these patient-specific residual models for continuous-valued clinical time-series and achieve better forecasting performance. 

\noindent \textbf{Online Switching Methods.} The online switching (selection) method is a complementary approach that has been used to increase prediction performance of online personalization models by allowing multiple (candidate) models to be used together \cite{littlestone1994weighted,shalev2011online}.
At each time in a sequential process, a switching decision is made based on recent prediction performance of each candidate model. For example, for continuous-valued clinical time-series prediction, Liu and Hauskrecht \cite{liu2017personalized} have a pool of population and patient-specific time-series models and at any point of time the switching method selects the best performing model. 

\noindent \textbf{Neural Clinical Event Sequence Prediction.} EHR-derived clinical event sequence data consists of thousands of sparse 
and infrequently occurring 
clinical events. 
In recent years, neural-based models have become the most popular and also the most successful models for representing and predicting EHR-derived clinical sequence data. 
The advantages of such models are their flexibility in modeling latent structures, feature representation, and their learning capability. 
Specifically, word embedding methods 
\cite{mikolov2013distributed}
are effectively used to learn low-dimensional compact representation (embedding) of clinical concepts
\cite{choi2016multi} and predictive patient state representations \cite{tran2015learning}. 
For autoregressive event prediction task, hidden state-space models (e.g., RNN, GRU) and attention mechanism are applied to learn latent dynamics of patient states progression and predict clinical variables such as
diagnosis codes \cite{malakouti2019predicting,malakouti2019hierarchical},
ICU mortality risk \cite{yu2020monitoring}, 
heart failure onset \cite{choi2016retain},
and multivariate future clinical event occurrences \cite{lee2019context,lee_clinical_2020,lee2020multi,lee2021modeling,liu2019nonparametric}.
For neural-based personalized clinical event prediction, most works focus on using patient-specific feature embedding obtained from patient demographics features \cite{gao2019camp,zhang2018patient2vec}. 
A limitation of the approach is that complex transitions of patient states in time cannot be modeled in a personalized way through static feature embeddings.   
In this work, we develop and investigate methods for adapting modern autoregressive models based on RNN that have been successfully applied to various complex clinical patient states and prediction models.  

\vspace{-3mm}
\section{Methodology}
\vspace{-3mm}
\subsection{Neural Autoregressive Event Sequence Prediction}
\label{simple-rnn}
\vspace{-3mm}
Our goal is to predict occurrences of multiple target events in clinical event sequences. We aim to build an autoregressive model $\phi$ that can predict, at any time $t$, the next step (target) event vector $\bm{y'}_{t+1}$ from a history of past (input) event vectors $\bm{H}_t = \{\bm{y}_{1},\dotsc,\bm{y}_{t}\}$, that is, $\bm{\hat{y}}'_{t+1} = \phi(\bm{H}_t)$. The event vectors are binary $\{0,1\}$ vectors, one dimension per an event type. The input vectors are of dimension $|E|$ where $E$ are different event types in clinical sequences.  The target vector is of dimension $|E'|$, where $E'\subset E$ are events we are interested in predicting. 

One way to build a neural autoregressive prediction model $\phi$ is to use Recurrent Neural Network (RNN) with input embedding matrix $\bm{W}_{emb}$, output linear projection matrix $\bm{W}_{o}$, bias vector $\bm{b}_{o}$, and sigmoid (logit) activation function $\sigma$. At each time step $t$, the RNN-based autoregressive model $\phi$ reads new input $\bm{y}_t$, updates hidden state $\bm{h}_t$, and generates prediction of the target vector $\bm{\hat{y}}'_{t+1}$:

\vspace{-2mm}
\[ \begin{array}{lll}%
\bm{v}_t = \bm{W}_{emb} \cdot \bm{y}_t &\qquad 
\bm{h}_{t} = \text{RNN}(\bm{h}_{t-1}, \bm{v}_t) &\qquad 
\bm{\hat{y}}'_{t+1} = \sigma(\bm{W}_{o} \cdot \bm{h}_{t} + \bm{b}_{o})\\
\end{array}\]%
\vspace{-6mm}

$\bm{W}_{emb}, \bm{W}_{o}, \bm{b}_{o}$, and RNN's parameters are learned through SGD with loss function $\mathcal{L}$ defined by the binary cross entropy (BCE):
\vspace{-4mm}
\begin{equation}
\begin{aligned}
\label{loss}
\mathcal{L} &= \sum_{s \in \mathcal{D}} \sum_{t=1}^{T(s)-1} e(\bm{y'}_{t+1},\bm{\hat{y}'}_{t+1}) \\
\end{aligned}
\end{equation}
\vspace{-4mm}
\begin{equation}
\begin{aligned}
\label{bce-loss}
e(\bm{y'}_t,\bm{\hat{y}}'_t) &= - [\bm{y'}_{t} \cdot \log \bm{\hat{y}}'_{t} + (\bm{1} - \bm{y'}_{t}) \cdot \log (\bm{1} - \bm{\hat{y}}'_{t})]
\vspace{-4mm}
\end{aligned}
\end{equation}
where $\mathcal{D}$ is training set and $T(s)$ is length of a sequence $s$. 
This neural autoregressive approach has several benefits when modeling complex high-dimensional clinical sequences: First, low-dimensional embedding with $\bm{W}_{emb}$ helps us to obtain a compact representation of high-dimensional input vector $\bm{y}$. Second, complex dynamics of observed patient state sequences are modeled through RNN which is capable of modeling non-linearities of the sequences. Furthermore, latent variables of neural models typically do not assume a specific probability form. Instead, the complex input-output association is learned through SGD based end-to-end learning framework which allows more flexibility in modeling complex latent dynamics of observed sequence.

However, the neural autoregressive approach cannot address one important characteristic of the clinical sequence: the variability in the dynamics of sequences across different patients. Typically, EHR-derived clinical sequences consist of medical history of several tens of thousands of patients. The dynamics of one patient's sequence could be significantly different from the sequences of other patients. For typical neural autoregressive models, parameters of the trained model are used to process and predict sequences of \textit{all} patients which consist of individual patients who can have different types of clinical complications, medication regimes, or observed sequence dynamics.

\vspace{-7mm}

\begin{algorithm}
    \SetKwInOut{Input}{Input}
    \SetKwInOut{Output}{Output}
\SetAlgoLined
\caption{Online Model Adaptation}
\label{algo:online-update}
\Input{Population model $\phi^{P}$, Current patient's history of \textbf{observed} input sequence $\bm{H}_t=\{\bm{y}_{1},\dotsc,\bm{y}_{t}\}$ and target sequence $(\bm{y'}_{1},\dotsc,\bm{y'}_{t})$}

Initialize patient-specific model $\phi^{I}$ from $\phi^{P}$; $\tau=0$; $\mathcal{L}^{*}_t(0)=\infty$\;
\Repeat{$\mathcal{L}^{*}_t(\tau-1) - \mathcal{L}^{*}_t(\tau) < \epsilon$}
{
    $\tau = \tau + 1$\;
    $\mathcal{L}^{*}_t(\tau) = \sum_{i=1}^{t-1} e\big(\bm{y'}_{i+1},\bm{\hat{y}}'_{i+1}\big) \cdot K(t, i)$ where $\bm{\hat{y}}'_{i+1}=\phi^{I}(\bm{H}_i)$\; 
    
    Update parameters of $\phi^{I}$ with $\mathcal{L}^{*}_t(\tau)$ via SGD;
}
\Output{Patient-specific model $\phi^{I}$}
\end{algorithm}

\vspace{-12mm}

\subsection{Online Adaptation of Model Parameters} 
\vspace{-3mm}

To address the patient variability issue, we propose a novel learning framework that adapts the parameters of the neural autoregressive model to the current patient sequence via SGD. For simplicity, we denote population model $\phi^{P}$ as a model trained on all training set patient data and patient (instance)-specific model $\phi^{I}$ that adapted to the current patient sequence at the prediction (test) stage. 
As described in Algorithm \ref{algo:online-update}, the online model adaptation procedure at time $t$ for the current patient starts by creating a patient-specific model $\phi^{I}$ from the population model $\phi^{P}$. They have identical model architecture and values of parameters in $\phi^{I}$ are initialized from $\phi^{P}$. Then, we compute an online error $\mathcal{L}^{*}_t = \sum_{i=1}^{t-1} e(\bm{y'}_{i+1},\bm{\hat{y}}'_{i+1})K(t, i)$ that reflects how much the prediction of $\phi^{I}$ deviates from the already observed target sequence for the current patient. 
With $\mathcal{L}^{*}_t$, we iteratively update parameters of $\phi^{I}$ via SGD. Stopping criterion for the iterative update is: $\mathcal{L}^{*}_t(\tau-1) - \mathcal{L}^{*}_t(\tau) < \epsilon$ where $\tau$ denotes the epoch of adaptation update and $\epsilon$ is a positive threshold.

\noindent \textbf{Discounting.}
Please note that our adaptation-based loss $\mathcal{L}^{*}_t$ combines prediction errors for all time steps of the sequence. However, in order to better fit it to the most recent patient-specific behavior, it also biases the loss more towards recent clinical events. This is done by weighting prediction error for each step $i < t$ with $K(t, i)$ that is based on its time difference from the current time $t$. More specifically, $K(t, i)$ defines an exponential decay function: 
\vspace{-2mm}
\begin{equation}
\begin{aligned}
\label{eq:exp-kernel}
K(t, i) = \exp{\Big( - \frac{|t - i|}{\gamma} \Big)}
\end{aligned}
\vspace{-2mm}
\end{equation}
where $\gamma$ denotes the bandwidth (slope) of exponential decay; if $\gamma$ is close to $+\infty$, errors at all time steps have the same weight.

\noindent \textbf{Online Adaptation of Model Components.}
\label{sec:online-component}
The RNN model may have too many parameters, and it may not help to adapt to all of them at the same time. One solution is to relax and permit to adapt only a subset of parameters. We experiment with and compare the adaptation of output layer parameters ($\bm{W}_{o}, \bm{b}_{o}$) and transition model (RNN) parameters.

\vspace{-7mm} 

\begin{algorithm}
    \SetKwInOut{Input}{Input}
    \SetKwInOut{Output}{Output}
\SetAlgoLined
\Input{$\phi^{P}$, $\phi^{I}$, $\bm{H}_t = \{\bm{y}_{1},\dotsc,\bm{y}_{t}\}$,$(\bm{y'}_{1},\dotsc,\bm{y'}_{t})$} 
$\mathcal{L}^{I} = \sum_{i=1}^{t} e({\bm{y'}}_{i+1},{\bm{\hat{y'}}}_{i+1}^{I}) \cdot K(t, i)$ where $\bm{\hat{y'}}^{I}_{i+1}={\phi}^{I}(\bm{H}_i)$\; 
$\mathcal{L}^{P} = \sum_{i=1}^{t} e(\bm{y'}_{i+1},\bm{\hat{y'}}_{i+1}^{P} ) \cdot K(t, i)$ where $\bm{\hat{y'}}_{i+1}^{P}=\phi^{P}(\bm{H}_i)$\;

\eIf{${\mathcal{L}}^{P} \ge {\mathcal{L}}^{I}$}{
    $\bm{\hat{y'}}_{t+1} = \bm{\hat{y'}}_{t+1}^{I}$ 
}{
    $\bm{\hat{y'}}_{t+1} = \bm{\hat{y'}}_{t+1}^{P}$
}
\Output{Prediction at time step $t+1$: $\bm{\hat{y'}}_{t+1}$}
\caption{Online Model Switching}
\label{algo:online-switching}
\end{algorithm}

\vspace{-12mm}

\subsection{Adaptation by Model Switching}
\label{sec:online-switching}
\vspace{-3mm}
One limitation of online patient-specific adaptation is that it tries to modify the dynamics to fit more closely the specifics of the patient. However, when the patient state changes suddenly due to recent events (e.g., a sudden clinical complication such as sepsis), the parameters of the patient-specific model $\phi^{I}$ may not be able to adapt quickly enough to these changes. In such a case, switching back to a more general population model could be more desirable.   

Model switching framework \cite{liu2017personalized,shalev2011online} can resolve this issue by dynamically switching among a patient-specific model and the population model. Driven by the recent performance of models, it can switch to the best performing model at each time step. 
Algorithm \ref{algo:online-switching} implements the model switching idea. Given a trained population model $\phi^P$, a patient-specific model $\phi^I$ trained via online adaptation, and the current patient's observed sequence, we can compute discounted losses $\mathcal{L}^P, \mathcal{L}^I$ for both models on the past data. By comparing the two losses, we select the model that gives the best error and use it for predicting the next step.

\vspace{-5mm}
\section{Experimental Evaluation}
\vspace{-3mm}
\subsection{Experiment Setup}
\vspace{-2mm}
\subsubsection{Clinical Sequence Generation.}  We extract 5137 patients from publicly available MIMIC-3 database \cite{johnson2016mimic} using the following criteria: (1) age is between 18 and 99, (2) length of admission is between 48 and 480 hours, and (3) clinical records are stored in Meta Vision system, one of the systems used to create MIMIC-3. We generate train and test sets using 80/20 \% split ratio. 
From the extracted records, we generate multivariate event sequences with a sliding-window method. We segment all sequences with a time window $W$=$24$ hours. All events that occurred in a time-window are aggregated into a binary vector $\bm{y}_{i} \in \{0,1\}^{|E|}$ where $i$ denotes a time-step of the window and $E$ is a set of event types. At any point of time $t$, a sequence of vectors created from previous time-windows defines an (input) sequence. A vector representing events in the next time window defines the prediction target. 

\noindent \textbf{Feature Extraction.} We use medication administration, lab results, procedures, and physiological results to define events. For the first three categories, we remove events that were observed in less than 500 different patients. For physiological events, we select 16 important event types with the help of a critical care physician. Lab test results and physiological measurements with continuous values are discretized to high, normal, and low values based on normal ranges compiled by clinical experts. In terms of prediction targets, we only consider and represent events corresponding to occurrences of such events, and we do not predict their normal or abnormal values.  
This process results in 65 medications, 44 procedures, 155 lab tests, and 84 physiological events as prediction targets, for the total target vector size of 348. The input vectors are of size 449.

\noindent \textbf{Baseline Models.} We compare proposed models to the following baselines: 
\begin{itemize}
\vspace{-3mm}
\item \textbf{GRU-based POPulation model (GRU-POP)}: For RNN-based time-series modeling described in \Cref{simple-rnn}, we use GRU. 
($\lambda=$1e-05) The patient (INstance)-specific model (\textbf{GRU-IN}) has the same architecture.
\item \textbf{REverse-Time AttenTioN (RETAIN)}: RETAIN is a representative work on using attention mechanism to summarize clinical event sequences, proposed by Choi et al. \cite{choi2016retain}. It uses two attention mechanisms to comprehend the history of GRU-based hidden states in reverse-time order. For multi-label output, we use a sigmoid function at the output layer.  ($\lambda=$1e-05)
\item \textbf{Logistic regression based on Convolutional Neural Network (CNN)}: 
This model uses CNN to build predictive features summarizing the event history of patients. Following Nguyen et al. \cite{nguyen2016mathtt}, we implement this CNN-based model with a 1-dimensional convolution kernel followed by ReLU activation and max-pooling operation. 
To give more flexibility to the convolution operation, we use multiple kernels with different sizes (2,4,8) and features from these kernels are merged at a fully-connected (FC) layer. ($\lambda=$1e-05)
\vspace{-3mm}
\end{itemize}

\noindent \textbf{Model Parameters.} 
We use embedding dimension $64$, hidden state dimension $512$, for all neural models. The population model, RETAIN, and CNN use learning rate  $0.005$ and patient-specific models use $0.005$. To prevent over-fitting, we use L2 weight decay regularization during the training of GRU-POP, RETAIN, and CNN, and the weight $\lambda$ is determined by the internal cross-validation set (range: 1e-04, 1e-05, 1e-06, 1e-07). For the SGD optimizer, we use Adam.
For the early stopping criteria parameter, we set $\epsilon$=1e-04. 
For $\gamma$, we use fixed value 3.0.

\noindent \textbf{Evaluation Metric.} We use the area under the precision-recall curve (AUPRC) as the main evaluation metric. AUPRC is known for presenting a more accurate assessment of performance of models for a highly imbalanced dataset \cite{saito2015precision}.

\vspace{-5.2mm}
\subsection{Results on Online Adaptation vs. Population Model}
\begin{figure}[t]
    \vspace{-4mm}
    \centering
    \begin{minipage}{0.475\textwidth}
        \centering
        \centerline{\hspace{-4.5mm}\includegraphics[scale=0.39]{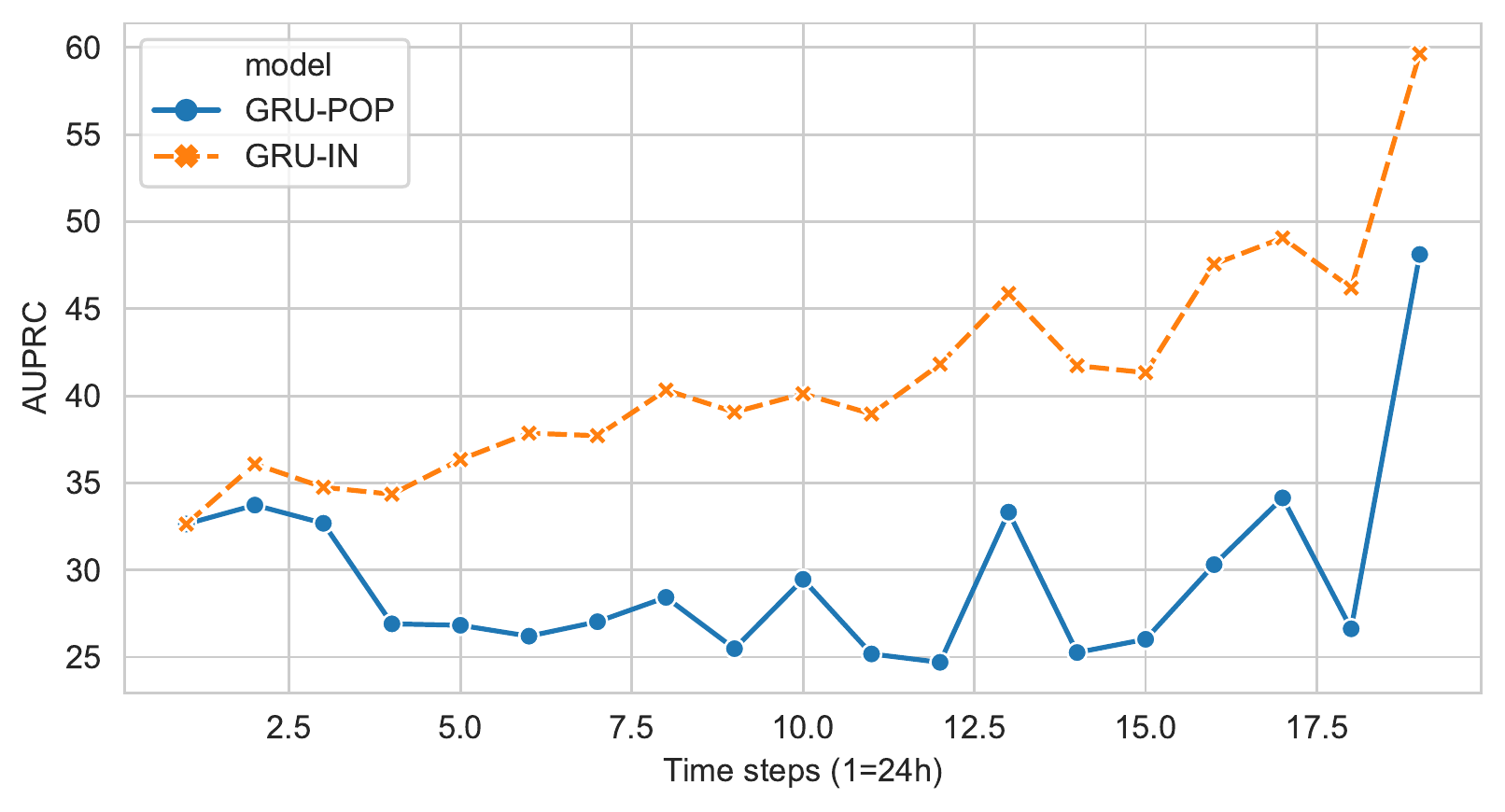}}
        \vspace*{-0.45cm}
        \caption{Prediction performance (AUPRC) of online adaptation method (GRU-IN) and population-based model (GRU-POP).}
        \label{figure:online adaptation}
    \end{minipage}\hfill
    \begin{minipage}{0.475\textwidth}
        \centering
        \centerline{\hspace{-4.5mm}\includegraphics[scale=0.39]{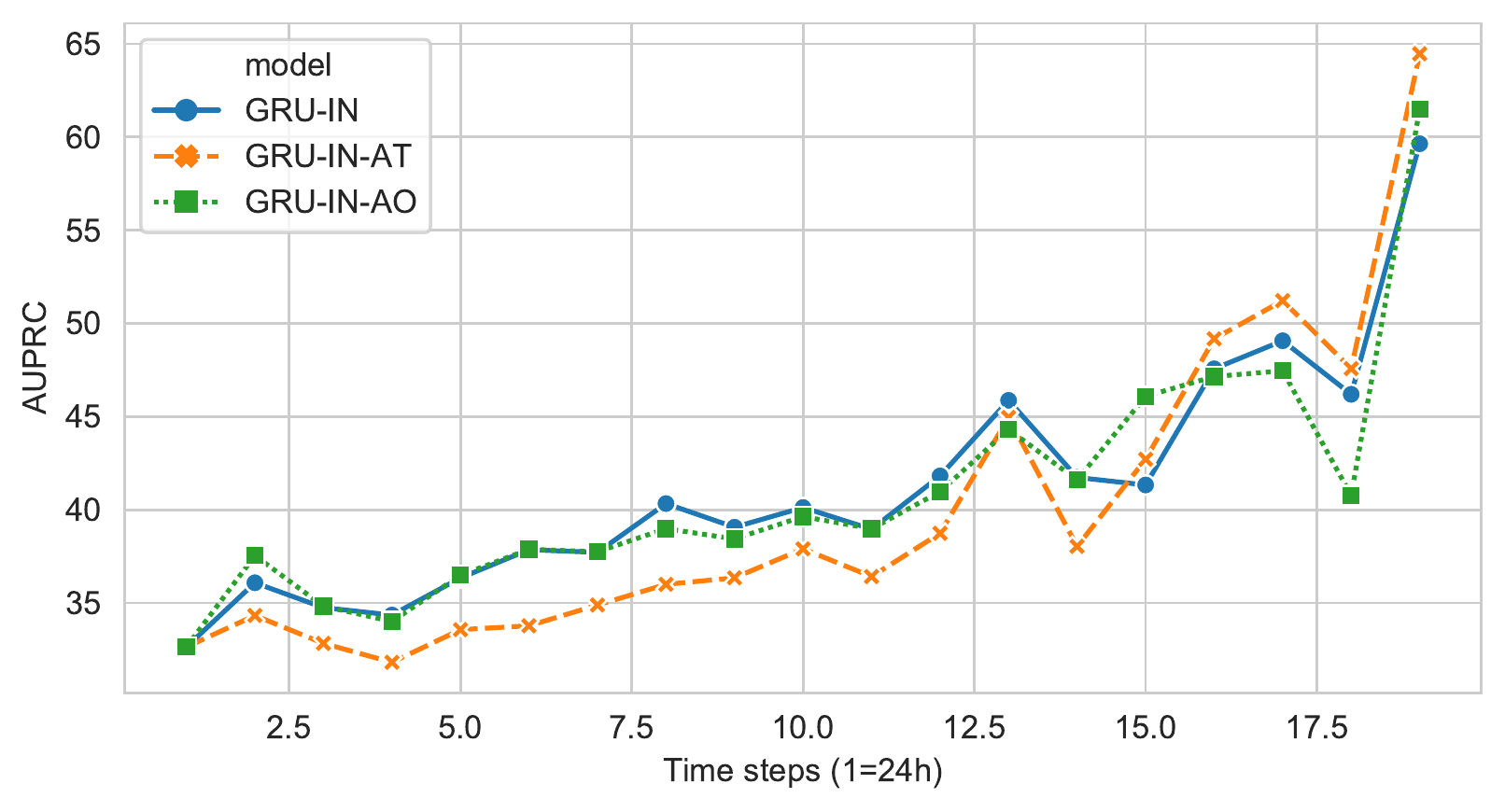}} 
        \vspace*{-0.45cm}
        \caption{Performance of online adaptation methods on all parameters (GRU-IN) and two subsets of parameters (GRU-IN-AT, GRU-IN-AO).}
        \label{figure:online component}
    \end{minipage}
    \vspace{-6mm}
\end{figure}

\vspace{-3mm}
We first compare the prediction performance of the population model (GRU-POP) and the proposed method on a patient-specific online adaption model that adapted all parameters (GRU-IN) as described in Algorithm \ref{algo:online-update}. 
As shown in \Cref{figure:online adaptation}, patient-specific model clearly outperforms population-based model across all time-steps. Especially in earlier days of admissions (day=1-3), the performance gap is smaller, but as time progresses on, the gap is increasing. It shows patient-specific online adaptation models can learn to more accurately predict patient-specific dynamics of event sequences compared to the population-based model.

\vspace{-5.2mm}
\subsection{Results on Adaptation on Partial Components}
\vspace{-3mm}
Next, we relax the online adaptation procedure to update only subsets of parameters: GRU-IN-AO is only adapting the output weight layer ($\bm{W}_{o},\bm{b}_{o}$) and GRU-IN-AT is only adapting parameters of the transition layer (GRU) only. As shown in \Cref{figure:online component}, the overall performance of GRU-IN-AO is close to GRU-IN which adapts all parameters. Since GRU-IN-AO is more efficient, it offers the best overall approach for patient-specific model adaptation. 

\vspace{-6mm}
\begin{table}
\centering
\small{
\begin{tabular}{lrrrrrr} 
\toprule
 & CNN & RETAIN & GRU-POP & GRU-IN & GRU-IN-SW & GRU-IN-AO-SW \\ \midrule
AUPRC & 30.81 & 29.67 & 29.61 & 41.13 & 42.14 & \underline{42.62} \\ \bottomrule
\vspace{-3mm}
\end{tabular}
}
\caption{Prediction results of all models averaged across all time steps} 
\vspace{-0.6cm}
\label{table:result-switching}
\end{table}

\vspace{-11mm}
\subsection{Results for Online Switching-based Adaptation} 
\begin{figure}[t]
\begin{center}
\vspace*{-4mm}
\centerline{\includegraphics[scale=0.42]{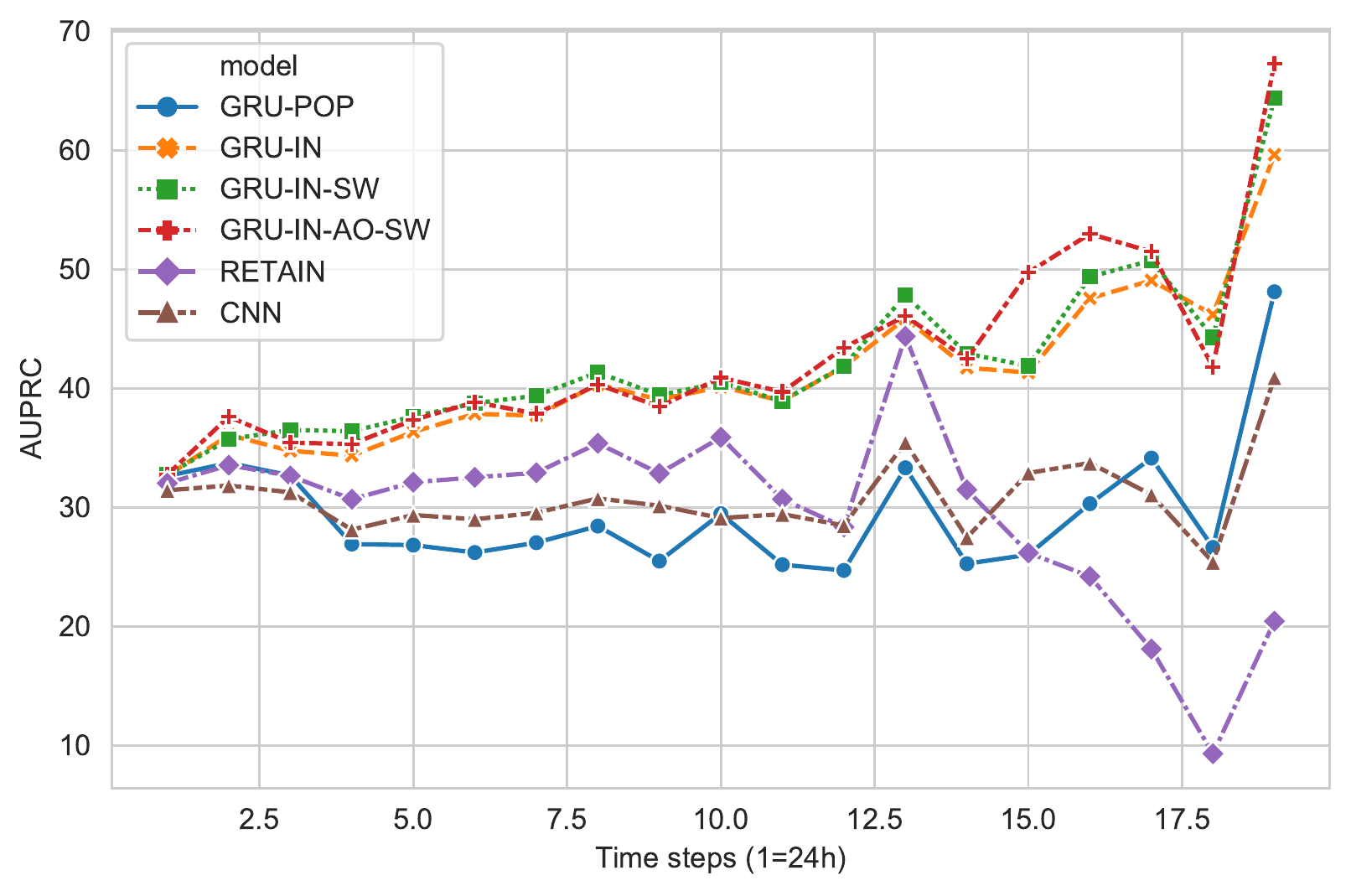}}
\vspace*{-4mm}
\caption{Performance of online switching methods (-SW) with population and patient-specific adaptation models.
Online switching methods 
clearly outperform baseline models 
(GRU-POP, RETAIN, CNN)
}
\vspace*{-12mm}
\label{figure:online switching}
\end{center}
\end{figure}

\vspace{-3mm}
We also experiment with online switching-based adaptation approach. It chooses the best predictive model from among a pool of available prediction models. We run the method to choose between a population-based model and a patient-specific adaptation model. 
We try the switching model in combination with the population model and two patient-specific models GRU-IN and GRU-IN-AO. 
The switching model results use post-fix '-SW'.
As shown in \Cref{figure:online switching}, models that rely on multiple models and online switching outperform baseline models of GRU-POP, CNN, and RETAIN. When the prediction performance is averaged across all time steps, we can observe that GRU-IN-AO-SW outperforms all models as shown in \Cref{table:result-switching}. Particularly, GRU-IN-AO-SW's AUPRC is +43\% higher than GRU-POP and RETAIN models. Compared to GRU-IN-AO (averaged AUPRC: 40.89), the online switching adaptation method increases AUPRC by +4.2\% and this reveals the benefit added by the online switching method.

\vspace{-6mm}
\subsubsection{When the model switches?}
\begin{figure}[t]
    \vspace{-4mm}
    \centering
    \begin{minipage}{0.475\textwidth}
        \centering
        \centerline{\hspace{-2.5mm}\includegraphics[scale=0.43]{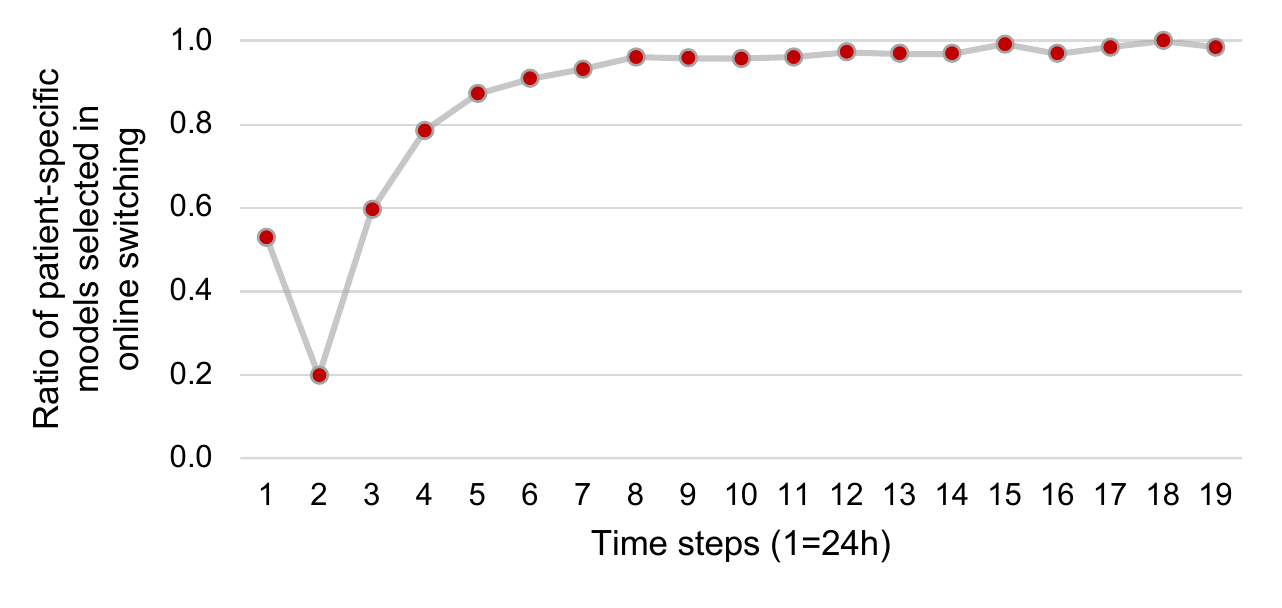}}
        \vspace*{-5mm}
        \caption{Ratio of patient-specific models selected in GRU-IN-SW. On latter time, online switching mechanism choose personalized models.}    
        \label{figure:ratio-patient-specific-models}
    \end{minipage}\hfill
    \begin{minipage}{0.475\textwidth}
        \centering
        \centerline{\includegraphics[scale=0.41]{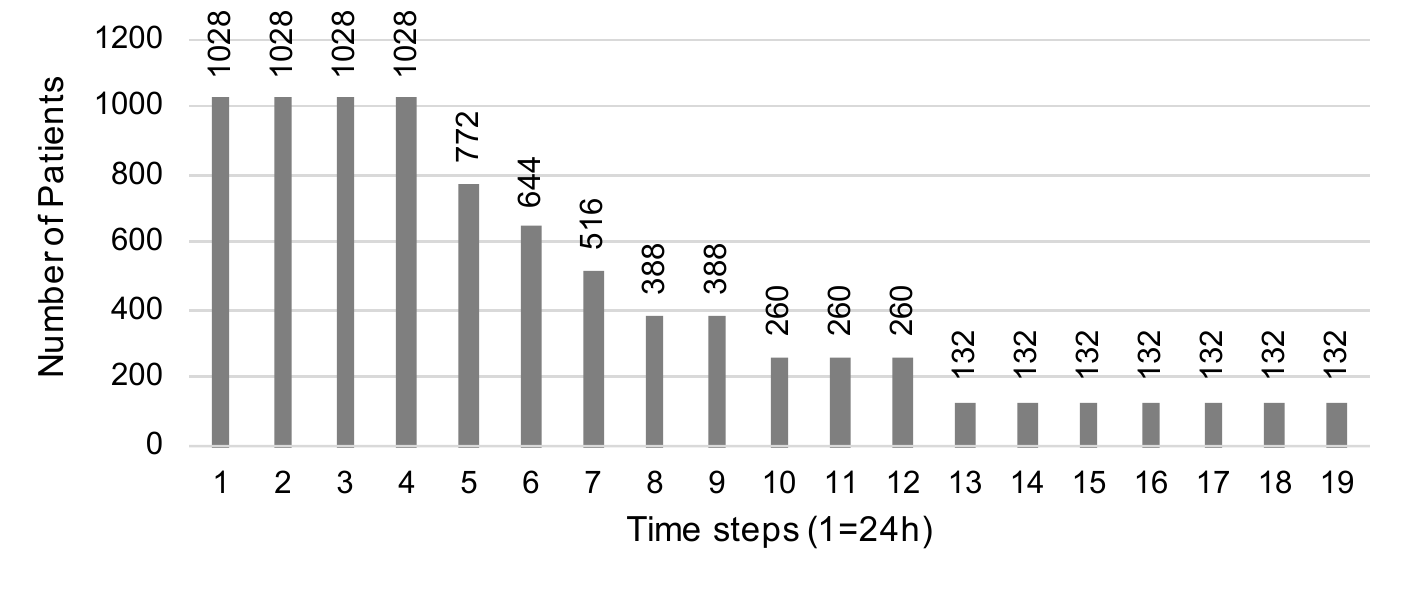}} 
        \vspace*{-5mm}
        \caption{Number of patients in each time step. The number of patients quickly deteriorates with longer sequence lengths.}        \label{figure:number-of-patients}
    \end{minipage}
\end{figure}

To have a better understanding of the behavior of online switching-based adaptation, we investigate when the model switches to a patient-specific model and to the population model.
First, we analyze how many times the online switching mechanism selects a patient-specific model (instead of a population model) over time and report the ratio of it. As shown in \Cref{figure:ratio-patient-specific-models}, in the early time steps, the online switching mechanism chooses the population model. However, at later time steps, the switching mechanism selects patient-specific models. This can be explained by the fact that patient-specific models need enough observations to adapt the patient-specific variability which is not possible with shorter sequences. 
To properly interpret the results, \Cref{figure:number-of-patients} shows the number of patients in each time step. This number can also be interpreted as the length of patient sequences and their volume. We can clearly see that the number of patients with longer sequences is very small, as the majority of sequences are very short. For example, patients with sequences longer than 13 days of admission are only about 12\% of all patients in test set. From this, we can conclude that the population model is often biased towards the dynamics and characteristics of shorter patient sequences. Meanwhile, patient-specific models can effectively learn and adapt better to the dynamics of longer sequences. 

\vspace*{-6mm}
\subsubsection{Predicting Repetitive and Non-Repetitive Events.} 
To perform this analysis, we divide event occurrences into two groups based on whether the same type of event has or has not occurred before. We compute AUPRC for each group as shown in \Cref{table:repeat-nonrepeat}.
The results show that for non-repetitive events, the performance of the patient-specific model is the lowest among all models. This is expected because with no previous occurrence of a target event, a patient-specific model could have difficulty making an accurate prediction for the new target event. In this case, we can also see the benefit of the online switching mechanism: the prediction of the population model is more accurate than the patient-specific model, and the online switching mechanism correctly chooses the population model. More specifically, GRU-IN-SW recovers most of the predictability of GRU-POP for non-repetitive event prediction. 
For repetitive event prediction, we can see that the patient-specific model (GRU-IN) outperforms the population-based models. However, the online switching approaches (GRU-IN-SW, GRU-IN-AO-SW) are the best and outperform all other approaches.

\begin{table}[t]
\vspace{-0.2cm}
\centering
\small{
\begin{tabular}{lrrrrrr} 
\toprule
 & CNN & RETAIN & GRU-POP & GRU-IN & GRU-IN-SW & GRU-IN-AO-SW \\ \midrule
Non-repetitive & 14.13 & 15.54 & \underline{15.85} & 11.11 & 15.55 & 14.37 \\ \hline
Repetitive & 45.16 & 50.30 & 52.04 & 52.83 & 53.73 & \underline{53.91} \\ \bottomrule
\end{tabular}
}
\caption{Prediction result on non-repetitive and repetitive event groups.
For non-repetitive events, the performance of patient-specific models (GRU-IN) is the lowest. The online switching approaches (GRU-IN-SW, GRU-IN-AO-SW) recover the predictability by switching to the population model.} 
\vspace{-9.5mm}
\label{table:repeat-nonrepeat}
\end{table}

\vspace{-0.4cm}
\section{Conclusion}
\vspace{-0.4cm}
In this work, we have developed methods for patient-specific adaptation of predictive models of clinical event sequences.  These models are of a great importance for defining representations of a patient state and for improving care. We demonstrate the improved performance of our models through experiments on MIMIC-3, a publicly available dataset of electronic health records for ICU patients. 

\vspace{-0.4cm}
\section*{Acknowledgement} 
\vspace{-0.4cm}
The work presented was supported by NIH grant R01GM088224. The content of this paper is solely the responsibility of the authors and does not necessarily represent the official views of NIH.
 
\vspace{-2mm}
\bibliographystyle{plain} 
\bibliography{BibFile-Shorter}

\begin{thebibliography}{10}

\bibitem{berzuini1992bayesian}
Carlo Berzuini et~al.
\newblock Bayesian networks for patient monitoring.
\newblock {\em Artificial Intelligence in Medicine}, 4:243–260, 05 1992.

\bibitem{choi2016multi}
Edward Choi et~al.
\newblock Multi-layer representation learning for medical concepts.
\newblock In {\em The 22nd ACM SIGKDD}, 2016.

\bibitem{choi2016retain}
Edward Choi et~al.
\newblock {RETAIN}: An interpretable predictive model for healthcare using
  reverse time attention mechanism.
\newblock In {\em Advances in NeurIPS}, 2016.

\bibitem{fojo2017precision}
Anthony~T Fojo et~al.
\newblock A precision medicine approach for psychiatric disease based on
  repeated symptom scores.
\newblock {\em Journal of psychiatric research}, 2017.

\bibitem{gao2019camp}
Jingyue Gao et~al.
\newblock {CAMP}: Co-attention memory networks for diagnosis prediction in
  healthcare.
\newblock ICDM, 2019.

\bibitem{hauskrecht2016outlier}
Milos Hauskrecht et~al.
\newblock Outlier-based detection of unusual patient-management actions: an icu
  study.
\newblock {\em Journal of biomedical informatics}, 64:211--221, 2016.

\bibitem{huang2015medical}
Zhengxing Huang et~al.
\newblock Medical inpatient journey modeling and clustering: a {Bayesian}
  hidden {Markov} model based approach.
\newblock In {\em AMIA}, volume 2015.

\bibitem{huang2013similarity}
Zhengxing Huang et~al.
\newblock Similarity measure between patient traces for clinical pathway
  analysis: problem, method, and applications.
\newblock {\em IEEE J-BHI}, 2013.

\bibitem{johnson2016mimic}
Alistair~EW Johnson et~al.
\newblock {MIMIC-III}, a freely accessible critical care database.
\newblock {\em Scientific data}, 3:160035, 2016.

\bibitem{lee2019context}
Jeong~Min Lee and Milos Hauskrecht.
\newblock Recent context-aware {LSTM}-based clinical time-series prediction.
\newblock In {\em Intl Conf on {AI} in {Medicine} ({AIME})}, 2019.

\bibitem{lee_clinical_2020}
Jeong~Min Lee and Milos Hauskrecht.
\newblock Clinical {Event} {Time}-series {Modeling} with {Periodic} {Events}.
\newblock In {\em The 33rd {International} {FLAIRS} {Conference}}, 2020.

\bibitem{lee2020multi}
Jeong~Min Lee and Milos Hauskrecht.
\newblock Multi-scale temporal memory for clinical event time-series
  prediction.
\newblock In {\em Intl Conf on {AI} in {Medicine} ({AIME})}, 2020.

\bibitem{lee2021modeling}
Jeong~Min Lee and Milos Hauskrecht.
\newblock Modeling multivariate clinical event time-series with recurrent
  temporal mechanisms.
\newblock {\em Artific. Intelligence in Medicine}, 2021.

\bibitem{littlestone1994weighted}
Nick Littlestone et~al.
\newblock The weighted majority algorithm.
\newblock {\em Inf. Comput.}, 108(2):212–261, February 1994.

\bibitem{liu2019nonparametric}
Siqi Liu and Milos Hauskrecht.
\newblock Nonparametric regressive point processes based on conditional
  gaussian processes.
\newblock In {\em Advances in NeurIPS}, 2019.

\bibitem{liu2016learning_a}
Zitao Liu and Milos Hauskrecht.
\newblock Learning adaptive forecasting models from irregularly sampled
  multivariate clinical data.
\newblock In {\em The 30th AAAI Conference}, 2016.

\bibitem{liu2017personalized}
Zitao Liu and Milos Hauskrecht.
\newblock A personalized predictive framework for multivariate clinical time
  series via adaptive model selection.
\newblock In {\em ACM CIKM}, 2017.

\bibitem{malakouti2019hierarchical}
Seyedsalim {Malakouti} and Milos Hauskrecht.
\newblock Hierarchical adaptive multi-task learning framework for patient
  diagnoses and diagnostic category classification.
\newblock In {\em IEEE BIBM}, 2019.

\bibitem{malakouti2019predicting}
Seyedsalim Malakouti and Milos Hauskrecht.
\newblock Predicting patient’s diagnoses and diagnostic categories from
  clinical-events in {EHR} data.
\newblock In {\em Intl Conf on {AI} in Medicine (AIME)}, 2019.

\bibitem{mikolov2013distributed}
Tomas Mikolov et~al.
\newblock Distributed representations of words and phrases and their
  compositionality.
\newblock In {\em Advances in NeurIPS}, pages 3111--3119, 2013.

\bibitem{nguyen2016mathtt}
Phuoc Nguyen et~al.
\newblock Deepr: a convolutional net for medical records.
\newblock {\em IEEE journal of biomedical and health informatics},
  21(1):22--30, 2016.

\bibitem{rizopoulos2011dynamic}
Dimitris Rizopoulos.
\newblock Dynamic predictions and prospective accuracy in joint models for
  longitudinal and time-to-event data.
\newblock {\em Biometrics}, 2011.

\bibitem{saito2015precision}
Takaya Saito and Marc Rehmsmeier.
\newblock The precision-recall plot is more informative than {ROC} plot when
  evaluating binary classifiers on imbalanced datasets.
\newblock {\em PloS One}, 2015.

\bibitem{shalev2011online}
Shai Shalev-Shwartz et~al.
\newblock Online learning and online convex optimization.
\newblock {\em Foundations and trends in Machine Learning}, 2011.

\bibitem{tran2015learning}
Truyen Tran et~al.
\newblock Learning vector representation of medical objects via {EMR}-driven
  nonnegative restricted {Boltzmann} machines.
\newblock {\em JBI}, 54:96--105, 2015.

\bibitem{visweswaran2005instance}
Shyam Visweswaran and Gregory~F Cooper.
\newblock Instance-specific bayesian model averaging for classification.
\newblock In {\em Advances in NeurIPS}, 2005.

\bibitem{yu2020monitoring}
Ke~Yu et~al.
\newblock Monitoring {ICU} mortality risk with a long short-term memory
  recurrent neural network.
\newblock In {\em Pac Symp Biocomput}. World Scientific, 2020.

\bibitem{zhang2018patient2vec}
Jinghe Zhang et~al.
\newblock Patient2vec: A personalized interpretable deep representation of the
  longitudinal electronic health record.
\newblock {\em IEEE Access}, 6, 2018.

\end{thebibliography}

\end{document}